# Ambiguity resolution in a reductionistic parser *

**Atro Voutilainen & Pasi Tapanainen**
Research Unit for Computational Linguistics
P.O. Box 4 (Keskuskatu 8)
FIN-00014 University of Helsinki
Finland

## Abstract

We are concerned with dependency-oriented morphosyntactic parsing of running text. While a parsing grammar should avoid introducing structurally unresolvable distinctions in order to optimise on the accuracy of the parser, it also is beneficial for the grammarian to have as expressive a structural representation available as possible. In a reductionistic parsing system this policy may result in considerable ambiguity in the input; however, even massive ambiguity can be tackled efficiently with an accurate parsing description and effective parsing technology.

## 1 Introduction

In this paper we are concerned with grammar-based surface-syntactic analysis of running text. Morphological and syntactic analysis is here based on the use of **tags** that express surface-syntactic relations between functional categories such as Subject, Modifier, Main verb etc.; consider the following simple example:

```
I       PRON      @SUBJECT
see     V PRES    @MAINVERB
a       ART       @DET>N
bird    N         @OBJECT
.       FULLSTOP
```

In this type of analysis, each word gets a morphosyntactic analysis[1].

The present work is closely connected with two parsing formalisms, **Constraint Grammar** [Karlsson, 1990; Karlsson *et al.*, 1991; Voutilainen *et al.*, 1992; Karlsson *et al.*, 1993] and **Finite-state syntax** as advocated by [Koskenniemi, 1990; Tapanainen, 1991; Koskenniemi *et al.*, 1992]. The Constraint Grammar parser of English is a sequential modular system that assigns a shallow surface-true dependency-oriented functional analysis on running text, annotating each word with morphological and syntactic tags. The finite-state parser assigns a similar type of analysis, but it operates on all levels of ambiguity[2] in parallel rather than sequentially, enabling the grammarian to refer to all levels of structural description in a single uniform rule component.

ENGCG, a wide-coverage English Constraint Grammar and lexicon, was written 1989–1992, and the system is currently available[3]. The Constraint Grammar framework was proposed by Fred Karlsson, and the English Constraint Grammar was developed by Atro Voutilainen (lexicon, morphological disambiguation), Juha Heikkilä (lexicon) and Arto Anttila (syntax). There are a few implementations

*This paper is published in *Proceedings of EACL'93*, (pp. 393–403, Utrecht). The development of ENGCG was supported by TEKES, the Finnish Technological Development Center, and a part of the work on Finite-state syntax has been supported by the Academy of Finland.

[1] It consists of a base form, a morphological reading – part-of-speech, inflectional and other morphosyntactic features – and a syntactic-functional tag, flanked by '@'.

[2] Morphological, clause boundary, and syntactic ambiguities

[3] The ENGCG parser can currently be tested automatically via E-mail by sending texts of up to 300 words to engcg@ling.Helsinki.FI. The reply will contain the analysis as well as information on usage and availability. Questions can also be directly sent to avoutila@ling.Helsinki.FI or to ptapanai@ling.Helsinki.FI.

of the parser, and the latest, written in C by Pasi Tapanainen, analyses more than 1000 words per second on a Sun SparcStation10, using a disambiguation grammar of some 1300 constraints.

Intensive work within the finite-state framework was started by Tapanainen [1991] in 1990, and an operational parser was in existence the year after. The first nontrivial finite-state descriptions [Koskenniemi *et al.*, 1992] were written by Voutilainen 1991–1992, and currently he is working on a comprehensive English grammar which is expected to reach a considerable degree of maturity by the end of 1994. Much of this emerging work is based on the ENGCG description, (e.g. the ENGTWOL lexicon is used as such); however, the design of the grammar has changed considerably, as will be seen below.

We have two main theses. Firstly, knowledge-based reductionistic grammatical analysis will be facilitated rather than hindered by the introduction of (new) linguistically motivated and structurally resolvable distinctions into the parsing scheme, although this policy will increase the amount of ambiguity in the parser's input. Secondly, the amount of ambiguity in the input does not predict the speed of analysis, so introduction of new ambiguities in the input is not necessarily something to be avoided.

Next, we present some observations about the ENGCG parser: the linguistic description would become more economic and accurate if all levels of structural description were available at the outset of reductionistic parsing (or disambiguation of alternative readings). In Section 3 we report on some early experiments with finite-state parsing. In Section 4 we sketch a more satisfactory functional dependency-oriented description. A more expressive representation implies more ambiguity in the input; in Section 5 it is shown, however, that even massive ambiguity need be no major problem for the parser.

## 2 Constraint Grammar of English

A large-scale description has been written within the Constraint Grammar (CG) framework. CG parsing consists of the following sequential modules:

- Preprocessing and morphological analysis
- Disambiguation of morphological (e.g. part-of-speech) ambiguities
- Mapping of syntactic functions onto morphological categories
- Disambiguation of syntactic functions

Here we shall be concerned only with disambiguation of morphological ambiguities – this module, along with the TWOL-style morphological description ENGTWOL, is the most mature part of the ENGCG system.

The morphological description is based on [Quirk *et al.*, 1985]. For each word, a base form, a part of speech as well as inflectional and also derivational tags are provided, e.g.

```
("<*i>"
   ("i" <*> ABBR NOM SG)
   ("i" <*> <NonMod> PRON PERS NOM SG1))
("<see>"
   ("see" <SVO> V SUBJUNCTIVE VFIN)
   ("see" <SVO> V IMP VFIN)
   ("see" <SVO> V INF)
   ("see" <SVO> V PRES -SG3 VFIN))
("<a>"
   ("a" <Indef> DET CENTRAL ART SG))
("<bird>"
   ("bird" <SV> V SUBJUNCTIVE VFIN)
   ("bird" <SV> V IMP VFIN)
   ("bird" <SV> V INF)
   ("bird" <SV> V PRES -SG3 VFIN)
   ("bird" N NOM SG))
("<$.>")
```

Ambiguities due to part of speech and minor categories are common in English – on an average, the ENGTWOL analyser furnishes each word with two readings. The task of the morphological disambiguator is certainly a nontrivial one.

The disambiguator uses a hand-written constraint grammar. Here, we will not go into the technicalities of the CG rule formalism; suffice it to say that each constraint – presently some 1,300 in all – expresses a partial paraphrase of some thirty more general grammar statements, typically in the form of negative restrictions. – For instance, a constraint might reject verb readings in an ambiguous morphological analysis as contextually illegitimate if the immediately preceding word is an unambiguous determiner. This can be regarded as a roundabout partial statement about the form of a noun phrase: a determiner is followed by a premodifier or a noun phrase head, so all morphological readings that cannot act as nominal heads or premodifiers are to be discarded.

Here is the disambiguated representation of the sentence:

```
("<*i>"
   ("i" <*> <NonMod> PRON PERS NOM SG1))
("<see>"
   ("see" <SVO> V PRES -SG3 VFIN))
("<a>"
   ("a" <Indef> DET CENTRAL ART SG))
("<bird>"
   ("bird" N NOM SG))
("<$.>")
```

Overall, the morphological disambiguator has a very attractive performance. While the best known competitors – typically based on statistical methods (see e.g. [Garside *et al.*, 1987; Church, 1988]) – make a misprediction about part of speech in up to 5% of all words, the ENGCG disambiguator makes a false prediction only in up to 0.3% of all cases [Voutilainen, 1993]. So far, ENGCG has been used in a large-scale information management system (an ESPRIT II project called SIMPR: *Structured Information Management: Processing and Retrieval*). Currently ENGCG is also used for tagging the *Bank of English*, a 200-million word corpus established by the COBUILD team in Birmingham, England; the tagged corpus will become accessible to the research community.

What makes ENGCG interesting for the present discussion is the fact that the constraints are essentially partial expressions of the distribution of functional-syntactic categories. In other words, the generalisations underlying the disambiguation constraints pertain to a higher level of description than is explicitly coded in the input representation.

The high number and also the complexity of most of the constraints mainly results from the fact that direct reference to functional categories is not possible in the constraint grammar because syntactic functions are systematically introduced only after morphological disambiguation has become disactivated. Also explicit information about sentence-internal clause boundaries is missing, so a constraint, usually about clause-internal relations, has to ascertain that the words and features referred to are in the same clause – again in a roundabout and usually partial fashion.

Indeed, it is argued in [Voutilainen, 1993] that if direct reference to all appropriate categories were possible, most or all of part-of-speech disambiguation would be a mere side-effect of genuine functional-syntactic analysis. In other words, it seems that the availability of a more expressive grammatical representation would make part-of-speech analysis easier, even though the amount of ambiguity would increase at the outset.

The ENGCG disambiguator avoids risky predictions; some 3–6% of all words remain partly ambiguous after part-of-speech disambiguation. Also most of these remaining ambiguities appear structurally resolvable. The reason why these ambiguities are not resolved by the ENGCG disambiguator is that the expression of the pertinent grammar rules as constraints, without direct reference to syntactic-function labels and clause boundaries, becomes prohibitively difficult. Our hypothesis is that also most of the remaining part-of-speech ambiguities could be resolved if also clause boundary and syntactic descriptors were present in the input, even though this would imply more ambiguity at the outset of parsing.

## 3 First experiences with Finite-State syntax

Finite-state syntax, as originally proposed by Koskenniemi, is an emerging framework that has been used in lexicon-based reductionistic parsing. Some nontrivial English grammars of some 150–200 rules have been written recently. The main improvements are the following.

- All three types of structural ambiguity – morphological, clause boundary, and syntactic – are presented in parallel. No separate, potentially sequentially applied subgrammars for morphological disambiguation, clause boundary determination, or syntax proper, are needed – one uniform rule component will suffice for expressing the various aspects of the grammar. In this setting, therefore, a genuine test of the justification of three separate types of grammar is feasible: for instance, it is possible to test, whether morphological disambiguation is reducible to essentially syntactic-functional grammar.

- The internal representation of the sentence is more distinctive. The FS parser represents each sentence reading separately, whereas the CG parser only distinguishes between alternative word readings. Therefore the FS rules need not concern themselves with more than one unambiguous, though potentially unacceptable, sentence reading at a time, and this improves parsing accuracy.

- The rule formalism is more expressive and flexible than in CG; for instance, the full power of regular expressions is available. The most useful kind of rule appears to be the **implication rule**; consider the following (somewhat simplified) rule about the distribution of the subject in a finite clause:

```
Subject =>
            _ .. FinVerbChain,
FinAux .. _ .. NonFinMainVerb ... QUESTION;
```

It reads: 'A finite clause subject (a constant defined as a regular expression elsewhere in the grammar) occurs before a finite verb chain in the same clause ('..'), or it occurs between a finite auxiliary and a nonfinite main verb in the same clause, and the sentence ends in a question mark.' – If a sentence reading contains a sequence of tags that is accepted by the regular expression *Subject* and that is not legitimated by the contexts, the sentence reading is discarded; otherwise it survives the evaluation, perhaps to be discarded by some other grammar rule.

Implication rules express distributions in a straightforward, positive fashion, and usually they are very compact: several dozens of CG rules that express bits and pieces of the same grammatical phenomenon can usually be expressed with one or two transparent finite-state rules.

- The CG syntax was somewhat shallow. The difference between finite and non-finite clauses was mostly left implicit, and the functional description was not extended to clausal constructions, which also can serve e.g. as subjects and objects. In contrast, even the earlier FS grammars did distinguish between finite and non-finite constructions, although the functional description of these categories was still lacking in several respects. Still, even this modest enrichment of the grammatical representation made it easier to state distributional generalisations, al-though much still remained hard to express, e.g. coordination of formally different but functionally similar categories.

### 3.1 A pilot experiment

To test whether the addition of clause boundary and functional-syntactic information made morphological disambiguation easier, a finite-state grammar consisting of some 200 syntactic rules [Koskenniemi *et al.*, 1992] was written, and a test text[4] was selected. The objective was to see, whether those morphological ambiguities that are too hard for the ENGCG disambiguator to resolve can be resolved if a more expressive grammatical description (and a more powerful parsing formalism) is used.

Writing a text-generic comprehensive parsing grammar of a maturity comparable to the ENGCG description would have taken too much time to be practical for this pilot test. While most of the grammar rules were about relatively frequently occurring constructions, e.g. about the structure of the finite verb chain or of prepositional phrases, some of the rules were obviously 'inspired' by the test text: the test grammar is more comprehensive on the structural phenomena of the test text than on texts in general. However, *all* proposed rules were carefully tested against various corpora, e.g. a manually tagged collection of some 2,000 sentences taken from [Quirk *et al.*, 1985], as well as large untagged corpora, in order to ascertain the generality of the proposed rules.

Thus the resulting grammar was 'optimised' in the sense that all syntactic structures of the text were described in the grammar, but *not* in the sense that the rules would have been true of the test text only.

The test data was first analysed with the ENGCG disambiguator. Out of the 1,400 words, 43 remained ambiguous due to morphological category, and no misanalyses were made. Then the analysed data was enriched with the more expressive finite-state syntactic description, i.e. with new ambiguities, and this data was then analysed with the finite-state parser. After finite-state parsing, only 3 words remained morphologically ambiguous, with no misanalyses. Thus the introduction of more descriptive elements into the sentence representations made it possible to safely resolve almost all of the remaining 43 morphological ambiguities.

This experiment suggests the usefulness of having available as much structural information as possible, although undoubtedly some of the additional precision resulted from a more optimal internal representation of the input sentence and from a more expressive rule formalism. Overall, these results seem to contradict certain doubts voiced [Sampson, 1987; Church, 1992] about the usefulness of syntactic knowledge in e.g. part-of-speech disambiguation. Part-of-speech disambiguation is essentially syntactic in nature; at least current methods based on lexical probabilities provide a less reliable approximation of correct part-of-speech tagging.

## 4 A new tagging scheme

The above observations suggest that grammar-based analysis of running text is a viable enterprise – not only academically, but even for practical applications. A description that on the one hand avoids introducing systematic structurally unresolvable ambiguities, and, on the other, provides an expressive structural description, will, together with a careful and detailed lexicography and grammar-writing, make for a robust and very accurate parsing system.

The main remaining problem is the shortcomings in the expressiveness of the grammatical representation. The descriptions were somewhat too shallow for conveniently making functional generalisations at higher levels of abstraction; this holds especially for the functional description of non-finite and finite clauses.

This became clear also in connection with the experiment reported in the previous section: although the number of remaining morphological ambiguities was only three, the number of remaining *syntactic* ambiguities was considerably higher: of the 64 sentences, 48 (75%) received a single syntactic analysis, 13 sentences (20%) received two analyses, one sentence received three analyses, and two sentences received four analyses.

Here, we sketch a more satisfying notation that has already been manually applied on some 20,000 words of running text from various genres as well as on some 2,000 test sentences from a large grammar [Quirk *et al.*, 1985]. Together, these test corpora serve as a first approximation of the inventory of syntactic structures in written English, and they can be conveniently used in the validation of the new grammar under development.

### 4.1 Tags in outline

The following is a schematic representation of the syntactic tags:

| | |
|---|---|
| SUBJ | Subject |
| F-SUBJ | Formal subject |
| OBJ | Object |
| F-OBJ | Formal object |
| I-OBJ | Indirect object |
| SC | Subject complement |
| OC | Object complement |
| P<< | Preposition complement |
| >>P | Complement of deferred preposition |
| APP | Apposition |
| | |
| @>A | AD-A, head follows |
| @A< | AD-A, head precedes |

[4] An article from *The New Grolier Electronic Encyclopedia*, consisting of some 1,400 words

| | |
|---|---|
| @>N | Determiner or premodifier |
| @>P | Modifier of a PP |
| N< | Postdeterminer or postmodifier |
| ADVL | Adverbial |
| ADVL/N< | Adverbial or postmodifier |
| @CC | Coordinator |
| @CS | Subordinator |
| AUX | Auxiliary |
| MV | Main verb |
| MAINC | Main clause |
| mainc | Non-finite verbal fragment |
| n-head | Nominal fragment |
| a-head | Adverbial fragment |

This list represents the tags in a somewhat abstract fashion. Our description also employs a few notational conventions.

Firstly, the notation makes an explicit difference between two kinds of clause: the finite and the non-finite.

A finite clause typically contains (i) a verb chain, one or more in length, one of which is a finite verb, and (ii) a varying number of nominal and adverbial constructs. Verbs and nominal heads in a finite clause are indicated with a tag written in the upper case, e.g. *Sam/@SUBJ was/@MV a/@>N man/@SC.*

A verb chain in a non-finite clause, on the other hand, contains only non-finite verbs. Verbs and nominal heads in a non-finite clause are indicated with a tag written in the lower case, e.g. *To/@aux be/@mv or/@CC not/@ADVL to/@aux be/@mv.*

While a distinction is made between the upper and the lower case in the description of verbs and nominal heads, no such distinction is made in the description of other categories, which are all furnished with tags in the upper case, cf. *or/@CC not/@ADVL.*

Secondly, the notation accounts both for the internal structure of clausal units and for their function in their matrix clause. Usually, all tags start with the '@' sign, but those tags that indicate the function of a clausal unit rather than its internal structure *end* with the '@' sign. The function tag of a clause is attached to the main verb of the clause, so main verbs always get *two* tags instead of the ordinary one tag. An example is in order:

```
How     @ADVL
to      @aux
write   @mv      mainc@
books   @obj
.
```

Here *write* is a main verb in a non-finite clause (*@mv*), and the non-finite clause itself acts as an independent non-finite clause (*mainc@*).

### 4.2 Sample analyses

Next, we examine the tagging scheme with some concrete examples. Note, however, that most morphological tags are left out in these examples; only a part-of-speech tag is given. Consider the following analysis:

```
                                            @@
smoking     PCP1      @mv      SUBJ@         @
cigarettes  N         @obj                   @
inspires    V         @MV      MAINC@        @
the         DET       @>N                    @
fat         A         @>N                    @
butcher's   N         @>N                    @
wife        N         @OBJ                   @
and         CC        @CC                    @
daughters   N         @OBJ                   @
.           FULLSTOP                        @@
```

The boundary markers '@@', '@', '@/', '@<' and '@>' indicate a sentence boundary, a plain word boundary, an iterative clause boundary, the beginning, and the end, of a centre embedding, respectively.

As in ENGCG, also here all words get a function tag. *Smoking* is a main verb in a non-finite construction (hence the lower case tag *@mv*); *cigarette* is an object in a non-finite construction; *inspires* is a main verb in a finite construction (hence the upper case tag *@MV*), and so on.

Main verbs also get a second tag that indicates the function of the verbal construction. The non-finite verbal construction *Smoking cigarettes* is a subject in a finite clause, hence the tag *SUBJ@* for *Smoking.* The finite clause is a main clause, hence the tag *MAINC@* for *inspires*, the main verb of the finite clause.

The syntactic tags avoid telling what can be easily inferred from the context. For instance, the tag *@>N* indicates that the word is a determiner or a premodifier of a nominal. A more detailed classification can be achieved by consulting the morphological codes in the same morphological reading, so from the combination *DET @>N* we may deduce that *the* is a determiner of a nominal in the right-hand context; from the combination *A @>N* we may deduce that *fat* is an adjectival premodifier of a nominal, and so forth.

The notation avoids introducing structurally unresolvable distinctions. Consider the analysis of *fat.* The syntactic tag *@>N* indicates that the word is a premodifier of a nominal, and the head is to the right – either it is the nominal head of the noun phrase, or otherwise it is another nominal premodifier in between. In other words, the tag *@>N* accounts for both of the following bracketings:

```
[[fat butcher's] wife]
[[fat [butcher's wife]
```

Note also that coordination often introduces unresolvable ambiguities. On structural criteria, it is impossible to determine, for instance, whether *fat* modifies the coordinated *daughters* as well in *the fat butcher's wife and daughters*. Our notation keeps also this kind of ambiguity covert, which helps to keep the amount of ambiguity within reasonable limits.

In our description, the syntactic function is carried by the coordinates rather than by the coordinator – hence the object function tags on both *wife* and *daughters* rather than on *and*. An alternative convention would be the functional labelling of the conjunction. The difference appears to be merely notational.

A distinction is made between finite and non-finite constructions. As shown above, non-finiteness is expressed with lower case tags, and finite (and other) constructions are expressed with upper case tags. This kind of splitup makes the grammarian's task easier. For instance, the grammarian might wish to state that a finite clause contains maximally one potentially coordinated subject. Now if potential subjects in non-finite clauses could not be treated separately, it would be more difficult to express the grammar statement as a rule because extra checks for the existence of subjects of non-finite constructions would have to be incorporated in the rule as well, at a considerable cost to transparency and perhaps also to generality. Witness the following sample analysis:

```
                                      @@
Henry     N         @SUBJ              @
dislikes  V         @MV     MAINC@     @
her       PRON      @subj              @
leaving   PCP1      @mv     OBJ@       @
so        ADV       @>A                @
early     ADV       @ADVL              @
.         FULLSTOP                    @@
```

Apparently, there are two simplex subjects in the same clause; what makes them acceptable is that they have different verbal regents: *Henry* is a subject in a finite clause, with *dislikes* as the main verb, while *her* occurs in a non-finite clausal construction, with *leaving* as the main verb.

With regard to the description of *so early* in the above sentence, the present description makes no commitments as to whether the adverbial attaches to *dislikes* or *leaving* – in the notational system, there is no separate tag for adverbials in non-finite constructions. The resolution of adverbial attachment often is structurally unresolvable, so our description of these distinctions is rather shallow.

Also finite clauses can have a nominal functions. Consider the following sample.

```
                                        @@
What        PRON     @SUBJ               @
makes       V        @MV     SUBJ@       @
them        PRON     @OBJ                @
acceptable  A        @OC                 @/
is          V        @MV     MAINC@      @/
that        CS       @CS                 @
they        PRON     @SUBJ               @
have        V        @MV     SC@         @
different   A        @>N                 @
verbal      A        @>N                 @
regents     N        @OBJ                @
.           FULLSTOP                    @@
```

Here *What makes them acceptable* acts as a subject in a finite clause, and *that they have different verbal regents* acts as a subject complement. – Clauses in a dependent role are always subordinate clauses that typically have a more fixed word order than main clauses. Thus clause-function tags like *SC@* can also be used in fixing clause-internal structure.

Another advantage of the introduction of clause-function tags is that restricting the distribution of clauses becomes more straightforward. If, for instance, a clause is described as a postmodifying clause, then it has to follow something to postmodify; if a clause is described as a subject, then it should also have a predicate, and so on. More generally: previous grammars contained some rules explicitly about clause boundary markers, for instance:

```
  @/ =>
VFIN .. _ .. VFIN;
```

In contrast, the grammar currently under development contains no rules of this type. Clause boundary determination is likely to be reducible to functional syntax, much as is the case with morphological disambiguation. This new uniformity in the grammar is a consequence of the enrichment of the description with the functional account of clauses.

Also less frequent of 'basic' word orders can be conveniently accounted for with the present descriptive apparatus. For instance, in the following sentence there is a 'deferred' preposition; here the complement is to the left of the preposition.

```
                                                @@
What              PRON      @>>P                 @
are               V         @AUX                 @
you               PRON      @SUBJ                @
talking           PCP1      @MV     MAINC@       @
about <Deferred>  PREP      @ADVL                @
?                 QUESTION                      @@
```

Here *@>>P* for *What* indicates that a deferred preposition is to be found in the right-hand context, and the morphological feature *<Deferred>* indicates that *about* has no complement in the right-hand context: either the complement is to the left, as above, or it is missing altogether, as in

```
                                             @@
This            PRON     @SUBJ                @
is              V        @MV     MAINC@       @
the             DET      @>N                  @
house           N        @SC                  @/
she             PRON     @SUBJ                @
was             V        @AUX                 @
```

```
looking         PCP1     @MV    N<@       @
for <Deferred>  PREP     @ADVL            @
.               FULLSTOP                 @@
```

Ellipsis and coordination often co-occur. For instance, if finite clauses are coordinated, the verb is often left out from the non-first coordinates:

```
                                     @@
Pushkin    N        @SUBJ             @
was        V        @MV    MAINC@     @
Russia's   N        @>N               @
greatest   A        @>N               @
poet       N        @SC               @/
,          COMMA                      @
and        CC       @CC               @
Tolstoy    N        @SUBJ             @
her        PRON     @>N               @
greatest   A        @>N               @
novelist   N        @SC               @
.          FULLSTOP                  @@
```

Here, *and Tolstoy her greatest novelist* is granted a clause status, as indicated by the presence of the iterative clause boundary marker '@/'.

Note that clausal constructions without a main verb do not get a function tag because at present the clause function tag is attached to the main verb. If the ellipsis co-occurs with coordination, then the presence of the coordinator in the beginning of the elliptical construction (i.e. to the right of the iterative clause boundary marker '@/') may be a sufficient clue to the function tag: it is to the left, in the first coordinate.

Verbless constructions also occur in simplex constructions. Consider the following real-text example:

```
                                             @@
Providing       PCP1     @mv    ADVL@     @<
the             DET      @>N              @
pin             N        @SUBJ            @
has             V        @AUX             @
been            V        @AUX             @
fully           ADV      @ADVL            @
inserted        V        @MV    obj@      @
into            PREP     @ADVL            @
the             DET      @>N              @
connect         PCP1     @>N              @
rod             N        @P<<             @>
,               COMMA                     @
final           A        @>N              @
centralization  N        @SUBJ            @
can             V        @AUX             @
,               COMMA                     @
if              CS       @CS              @
necessary       A        @sc              @
,               COMMA                     @
be              V        @AUX             @
done            PCP2     @MV    MAINC@    @
on              PREP     @ADVL            @
a               DET      @>N              @
press           N        @P<<             @
using           PCP1     @mv    ADVL@     @
the             DET      @>N              @
support         N        @>N              @
stop            N        @>N              @
button          N        @obj             @
and             CC       @CC              @
driver          N        @obj             @
.               FULLSTOP                 @@
```

In the analysis of *if necessary*, there is a subject complement tag for *necessary*. Subject complements typically occur in clauses; clauses in general are assigned a syntactic function in our description; here, however, no such analysis is given due to the lack of a main verb. Nevertheless, in this type of verbless construction there is a lexical marker in the beginning: a subordinating conjunction or a *WH* word, and from this we can imply that the verbless construction functions as an adverbial.

An alternative strategy for dealing with the functional analysis of verbless constructions would be the assignment of clause-function tags also to nominal and adverbial heads. This would increase the amount of ambiguity at the outset, but on the other hand this new ambiguity would be easily controllable: a clausal construction serves only one function at a time in our description, and this restriction can be easily formalised in the finite-state grammar formalism.

Next, let us consider the description of prepositional phrases. In general, the present grammar tries to distinguish here between the adverbial function (*@ADVL*) and the postmodifier function (*@N<*). In the following somewhat contrived sentence, the distinction is straightforward to make in some cases.

```
                                        @@
Somebody    PRON @SUBJ                   @
with        PREP @N<                     @
a           DET  @>N                     @
telescope   N    @P<<                    @
saw         V    @MV    MAINC@           @
with        PREP @ADVL                   @
difficulty  N    @P<<                    @
the         DET  @>N                     @
man         N    @OBJ                    @
of          PREP @N<                     @
honor       N    @P<<                    @
with        PREP @ADVL/N<                @
the         DET  @>N                     @
binoculars  N    @P<<                    @
.           FULLSTOP                    @@
```

The phrase *with difficulty* is an unambiguous adverbial because it is directly preceded by a verb, which do not take postmodifiers. Likewise, *with a telescope* and *of honor* are unambiguously postmodifiers: the former because postnominal prepositional phrases without a verb in the left-hand context are postmodifiers; the latter because a postnominal *of*-phrase is always a postmodifier unless the left-hand context contains a member of a limited class of verbs like 'consist' and 'accuse' which take an *of*-phrase as a complement.

On the contrary, *with the binoculars* is a problem case: generally postnominal prepositional phrases with a verb in the left-hand context are ambiguous due to the postmodifier and adverbial functions. Furthermore, several such ambiguous prepositional phrases can occur in a clause at once, so in combination they can produce quite many grammatically acceptable analyses for a sentence. To avoid this uncomfortable situation, an underspecific tag has been introduced: a prepositional phrase is described unambiguously as *@ADVL/N<* if it occurs in a context legitimate for adverbials and postmodifiers – i.e., all other functions of prepositional phrases are disallowed in this context (with the exception of *of*-phrases). In all other contexts *@ADVL/N<* is disallowed.

This solution may appear clumsy, e.g. a new tag is introduced for the purpose, but its advantage is that description can take full benefit of the unambiguous 'easy' cases without paying the penalty of unmanageable ambiguity as a price for the extra information. – Overall, this kind of practise may be useful in the treatment of certain other ambiguities as well.

***

In this section we have examined the new tag scheme and how it responds to our two main requirements: the requirement of structural resolvability (cf. our treatment of premodifiers and prepositional phrases) and expressiveness of surface-syntactic relations (witness e.g. the manner in which the application of the Uniqueness principle as well as the description of clause distributions was made easier by extending the description).

It goes without saying that even the present annotation will leave some ambiguities structurally unresolvable. For instance, coordination is still likely to pose problems, cf. the following ambiguity due to the preposition complement and object analyses:

```
                                           @@
They          PRON  @SUBJ                   @
established   V     @MV   MAINC@            @
networks      N     @OBJ                    @
of            PREP  @N<                     @
state         N     @P<<                    @
and           CC    @CC                     @
local         A     @>N                     @
societies     N    [@OBJ --or-- @P<<]  @
.             FULLSTOP                     @@
```

Although the present system contains a powerful mechanism for expressing heuristic rules that can be used for ranking alternative analyses, the satisfactory treatment of ambiguities like this one seems to require some further adjustment of the tag scheme, e.g. further underspecification – something like our description of attachment ambiguities of prepositional phrases.

## 5 Ambiguity resolution with a finite-state parser

In a parsing system where all potential analyses are provided in the input to the parser, there is bound to be a considerable amount of ambiguity as the description becomes more distinctive. Consider the following sentence, 39 words in length:

```
A pressure lubrication system
is employed, the pump, driven
from the distributor shaft
extension, drawing oil from the
sump through a strainer and
distributing it through the
cartridge oil filter to a main
gallery in the cylinder block
casting.
```

If only part-of-speech ambiguities are presented, there are 10 million sentence readings. If each boundary between each word or punctuation mark is made four-ways ambiguous due to the word and clause boundary readings, the overall number of sentence readings gets as high as $10^{32}$ readings. If all syntactic ambiguities are added, the sentence representation contains $10^{66}$ sentence readings. Regarded in isolation, each word in the sentence is 1–70 ways ambiguous.

If we try to enumerate all $10^{66}$ readings and discard them one by one, the work is far too huge to be done. But we do not have to do it that way. Next we show that in fact the number of readings does not alone predict parsing complexity. We show that if we adopt a powerful rule formalism and an accurate grammar, which is also effectively applied, a lot of ambiguity can be resolved in a very short time.

We have seen above that very accurate analysis of running text can be achieved with a knowledge-based approach. Characteristic of such a system is the possibility to refer to grammatical categories at various levels of description within an arbitrarily long sentence context. – Regarding the viability of essentially statistical systems, the current experience is that employing a window of more than two or three words requires excessively hard computing. Another problem is that even acquiring collocation matrices based on e.g. four-grams or five-grams requires tagged corpora much larger than the current manually validated tagged ones are. Also, mispredictions, which are a very common problem for statistical analysers, tend to bring in the accumulation effect: more mispredictions are likely to occur at later stages of analysis. Therefore we do not have any reason to use unsure probabilistic information as long as we can use our more reliable linguistic knowledge.

Our rules can be considered as constraints that discard some illegitimate readings. When we apply rules one by one, the number of these readings decreases, and, if possible, in the end we have only one reading left. In addition to the ordinary 'absolute' rules, the grammar can also contain separate 'heuristic' rules, which can be used for ranking remaining multiple readings.

We represent sentences as finite state automata. This makes it possible to store all relevant sentence readings in a compact way. We also compile each grammar rule into a finite state automaton. Each rule automaton can be regarded as a constraint that accepts some readings and rejects some.

For example, consider the subject rule presented in Section 3. We can apply a rule like that on the sentence and, as a result, get an automaton that accepts all the sentence readings that are correct according to the rule. After this, our $10^{66}$–ways ambiguous sentence has, say, only some $10^{45}$ readings left. This means that in some fractions of a second, the number of readings is reduced into a 1/1000000000000000000000th part. All of these remaining readings are accepted by the applied rule. Next, we can apply another rule, and so on. The following rules will not probably reduce as many ambiguities as the first one, but they will reduce the ambiguity to some 'acceptable' level quite fast. This means that we cannot consider some sentences as unparsable just because they may initially contain a lot of ambiguity (say, $10^{100}$ sentence readings).

The real method we use is not as trivial as this, actually. The method presented above can rather be regarded as a declarative approach to applying the rules than as a description of a practical parser. A recent version of the parser combines several methods. First, it decreases the amount of ambiguity with some groups of carefully selected rules, as we described above. Then all other rules are applied together. This method seems [Tapanainen, 1992] to provide a faster parser than more straightforward methods.

Let us consider the different methods. In the first one we intersect a rule automaton with a sentence automaton and then we take another rule automaton that we intersect with the previous intermediate result, and so, on until all (relevant) rules have been applied. This method takes much time as we can see in the following table. The second method is like the first one but the rule automata have been ordered before processing: the most efficient rules are applied first. This ordering seems to make parsing faster. In the third method we process all rules together and the fourth method is the one that is suggested above. The last method is like the fourth one but also extra information is used to direct the parsing. It seems to be quite sufficient for parsing.

Before parsing commences, we can also use two methods for reducing the number of rule automata. Firstly, because the rules are represented as automata, a set of them can be easily combined using intersection of automata during the rule compilation phase. Secondly, typically not all rules are needed in parsing because the rule may be about some category that is not even present in the sentence. We have a quick method for selecting rules in run-time. These optimization techniques improve parsing times considerably.

Figure 1: Execution times of parsing methods (sec.).

| method | **1** | **2** | **3** | **4** | **5** |
|---|---|---|---|---|---|
| **non-opt.** | 31000 | 730 | 1500 | 500 | 290 |
| **optimized** | 7000 | 840 | 350 | 110 | 30 |

The test data is the same that was described above in Section 3.1. They were parsed on a Sun SparcStation 2.

The whole parsing scheme can be roughly presented as

- Preprocessing (text normalising and sentence boundary detection).
- Morphological analysis and enrichment with syntactic and clause boundary ambiguities.
- Transform each sentence into a finite state automaton.
- Select the relevant rules for the sentence.
- Intersect a couple of rule groups with the sentence automaton.
- Apply all remaining rules in parallel.
- Rank the resulting multiple analyses according to heuristic rules and select the best one if a totally unambiguous result is wanted.

## 6 Conclusion

It seems to us that it is the nature of the grammar rules, rather than the amount of the ambiguity itself, that determines the hardness of ambiguity resolution. It is quite easy to write a grammar that is extremely hard to apply even for simple sentence with a small amount of ambiguity. Therefore parsing problems that come up from using more or less incomplete grammars do not necessarily tell us about parsing text with a comprehensive grammar. Parsing problems due to ambiguity seem to dissolve if we have access to a more expressive grammatical representation; witness our experiences with morphological disambiguation using the two approaches discussed above.

We do not need to hesitate to use features that we consider useful in our grammatical description. The amount of ambiguity itself is not what enables or disables parsing. More important is that we have an effective grammar and parser that interact with each other in a sensible way, i.e. we should not try to kill mosquitos with artillery or to move mountains with a spoon. The ambiguity that is introduced has to be relevant for the grammar, not unmotivated or structurally unresolvable ambiguity, but ambiguity that provides us with information we need to resolve other ambiguities.

## References

[Church, 1988] Kenneth W. Church. A stochastic parts program and noun phrase parser for unrestricted text. In *Proceedings of the Second Conference on Applied Natural Language Processing*, pages 136–143, Austin, Texas, 1988.

[Church, 1992] Kenneth W. Church. Current Practice in Part of Speech Tagging and Suggestions for the Future. In Simmons (editor), *Abornik praci: In Honor of Henry Kučera, Michigan Slavic Studies*, pages 13–48, Michigan, 1992.

[Garside *et al.*, 1987] Garside, R., Leech, G. and Sampson, G., (editors) *The Computational Analysis of English. A Corpus-Based Approach.* Longman, London, 1987.

[Karlsson, 1990] Fred Karlsson. Constraint Grammar as a framework for parsing running text. In H. Karlgren (editor), *COLING-90. Papers presented to the 13th International Conference on Computational Linguistics.* Vol. 3 pages 168–173, Helsinki, 1990.

[Karlsson *et al.*, 1991] Karlsson, F., Voutilainen, A., Anttila, A. and Heikkilä, J. Constraint Grammar: a Language-Independent System for Parsing Unrestricted Text, with an Application to English. In *Natural Language Text Retrieval: Workshop Notes from the Ninth National Conference on Artificial Intelligence (AAAI-91).* Anaheim, California, 1991.

[Karlsson *et al.*, 1993] Karlsson, F., Voutilainen, A., Heikkilä, J. and Anttila, A. *Constraint Grammar: a Language-Independent System for Parsing Unrestricted Text.* (In print).

[Koskenniemi, 1990] Kimmo Koskenniemi. Finite-state parsing and disambiguation. In H. Karlgren (editor), *COLING-90. Papers presented to the 13th International Conference on Computational Linguistics.* Vol. 2 pages 229–232, Helsinki, 1990.

[Koskenniemi *et al.*, 1992] Kimmo Koskenniemi, Pasi Tapanainen and Atro Voutilainen. Compiling and using finite-state syntactic rules. In *Proceedings of the fifteenth International Conference on Computational Linguistics. COLING-92.* Vol. I pages 156–162, Nantes, France. 1992.

[Sampson, 1987] Geoffrey Sampson. Probabilistic Models of Analysis. In [Garside *et al.*, 1987].

[Tapanainen, 1991] Pasi Tapanainen. Äärellisinä automaatteina esitettyjen kielioppisääntöjen soveltaminen luonnollisen kielen jäsentäjässä (Natural language parsing with finite-state syntactic rules). Master's thesis. Dept. of computer science, University of Helsinki, 1991.

[Tapanainen, 1992] Pasi Tapanainen. Äärellisiin automaatteihin perustuva luonnollisen kielen jäsennin (A finite state parser of natural language). Licentiate (pre-doctoral) thesis. Dept. of computer science, University of Helsinki, 1992.

[Quirk *et al.*, 1985] Quirk, R., Greenbaum, S., Leech, G. and Svartvik, J. *A Comprehensive Grammar of the English Language.* Longman, London, 1985.

[Voutilainen, 1993] Atro Voutilainen. Morphological disambiguation. In [Karlsson *et al.*, 1993].

[Voutilainen *et al.*, 1992] Atro Voutilainen, Juha Heikkilä and Arto Anttila. *Constraint grammar of English. A Performance-Oriented Introduction.* Publications nr. 21, Dept. of General Linguistics, University of Helsinki, 1992.